\documentclass[letterpaper]{article} 
\usepackage{aaai23}  
\usepackage{times}  
\usepackage{helvet}  
\usepackage{courier}  
\usepackage[hyphens]{url}  
\usepackage{graphicx} 
\usepackage{natbib}  
\usepackage{caption} 
\usepackage{algorithm}
\usepackage{algorithmic}
\usepackage{subfiles}
\usepackage{xcolor}
\usepackage{adjustbox}
\usepackage{framed}
\usepackage{ragged2e}
\usepackage{blindtext}
\usepackage{amsmath}
\usepackage[normalem]{ulem}
\usepackage{enumitem}
\newlist{steps}{enumerate}{1}
\setlist[steps, 1]{label = \textbf{Step \arabic*:}}
\usepackage{newfloat}
\usepackage{listings}
\DeclareCaptionStyle{ruled}{labelfont=normalfont,labelsep=colon,strut=off} 
\floatstyle{ruled}
\newfloat{listing}{tb}{lst}{}
\floatname{listing}{Listing}
\title{Automating Knowledge Acquisition for Content-Centric Cognitive Agents Using LLMs}
\author{
   Sanjay Oruganti,
    Sergei Nirenburg\equalcontrib,
    Jesse English\equalcontrib,
    Marjorie McShane\equalcontrib
}
\affiliations{
    Cognitive Science Department, Rensselaer Polytechnic Institute, Troy, NY, 12180, USA\\
    zavedomo@gmail.com

 }

\usepackage{bibentry}

\begin{document}

\maketitle

\begin{abstract}
The paper describes a system that uses large language model (LLM) technology to support automatic learning of new entries in an intelligent agent’s semantic lexicon. The process is bootstrapped by an existing non-toy lexicon and a natural language generator that converts formal, ontologically-grounded representations of meaning into natural language sentences. The learning method involves a sequence of LLM requests and includes an automatic quality control step. To date, this learning method has been applied to learning multiword expressions whose meanings are equivalent to those of transitive verbs in the agent’s lexicon. The experiment demonstrates the benefits of a hybrid learning architecture that integrates knowledge-based methods and resources with both traditional data analytics and LLMs.
\end{abstract}
\section*{Introduction and Motivation}
Content-centric computational cognitive modeling \iftrue \cite{nirenburg2020content}  \fi stresses the importance of static knowledge resources as core components of cognitive architectures. To make cognitive agent systems deployable in real-world applications, they must be supported on vast amounts of knowledge about the world, about language, about themselves, and about other agents.  It is notoriously difficult to acquire such knowledge in amounts – and at the quality level – necessary to deploy AI members of human-AI teams performing real-world tasks. As the development of such agents is the main goal of our R\&D team, knowledge acquisition has been a central concern for a long time \cite{monarch1988ontos, viegas1995acquisition, viegas1996ecology, mcshane2021linguistics, nirenburg2023hybrid}. 
The general goal is to make knowledge acquisition less expensive over time by progressively automating it using any method or combination of methods that look promising. Our early efforts concentrated on data analytics support ergonomics of manual acquisition \iftrue \cite{wilks95steps} \fi. Once our team developed a reliable system for semantic and pragmatic analysis of text and acquired a non-toy set of knowledge resources to support it, we started experimenting with using the system itself to bootstrap an automatic learning process through learning by reading \iftrue\cite{nirenburg2007learning} \fi and dialog with a human \iftrue\cite{nirenburg2017toward} \fi. This paper reports on one of our first experiments on using large language models (LLMs) to support the next step in automating knowledge acquisition for cognitive agents.

Large Language Models (LLMs), based on the transformer architecture \cite{vaswani2017attention}, are crafted to emulate human-like responses by computing the best continuations (similar to dialog turns) for textual prompts provided by humans on the basis of training on vast amounts of stored text. LLMs excel in generating the next word in a sentence, and they neither understand what they are doing nor why. As a result, there are issues with deploying them in applications requiring trust based on explainability. Nevertheless, while LLM-oriented AI researchers are addressing this issue, LLMs can already be made useful as a tool, and the experiment we are describing is an example of such use.
\begin{figure}[!t]
    \centering
    \begin{framed}
    \includegraphics[trim= 70pt 290pt 285pt 75pt,clip, width=0.90\linewidth]{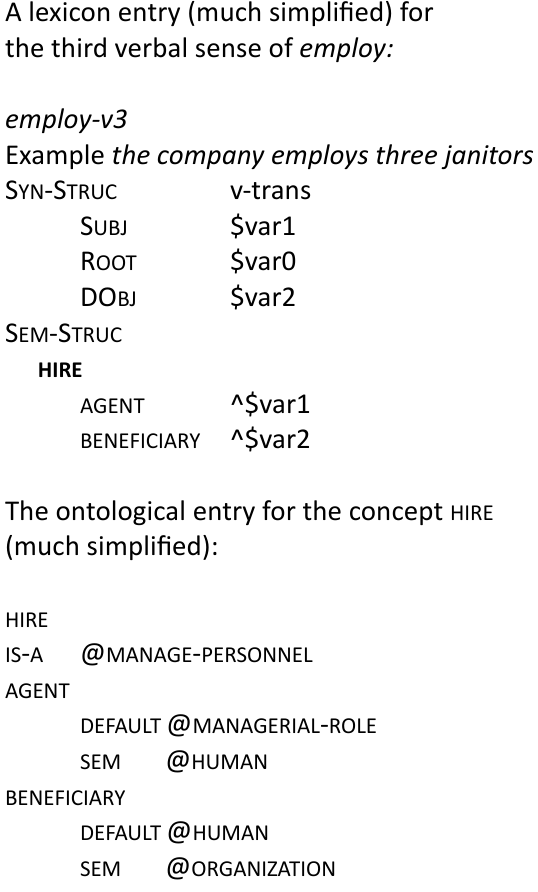}
          \end{framed}
    \caption{A sample lexicon entry and ontological concept}
    \label{fig:samplelexicon-onto}
    \vspace{-5pt}
\end{figure}
The experiment targets the automatic expansion of our architecture’s semantic lexicon that links words and constructions in natural language with their semantic and pragmatic meanings represented as elements of a formal ontology \iftrue\cite{mcshane2021linguistics}\fi.\footnote{In reality, lexicon entries contain more information, including a variety of dynamic meaning procedures that must be run to determine meanings of particular lexical units in context.} Our current English lexicon contains about 30,000 words and construction senses that are interpreted using an ontology of about 9,000 concepts. Each concept is characterized by an average of 16 properties. The present ontology uses about 350 of these properties that have the status of axioms. 

The experiment was devoted to learning the senses of verbs. A typical lexicon entry for a verb is illustrated in Figure \ref{fig:samplelexicon-onto}. The \textsc{sem-struc} zone of the entry refers to the underlying ontological concept and provides constraints on the verb’s case roles – it's \textsc{agent}, \textsc{theme}, \textsc{beneficiary}, etc., as appropriate. Additionally, Figure \ref{fig:samplelexicon-onto} also presents the ontological entry for \textsc{hire}, the concept supplying the basic meaning of \textit{employ-v3}. 
\begin{figure}[h]
    \centering
    \begin{framed}
        \includegraphics[trim= 70pt 450pt 300pt 75pt,clip, width=0.85\linewidth]{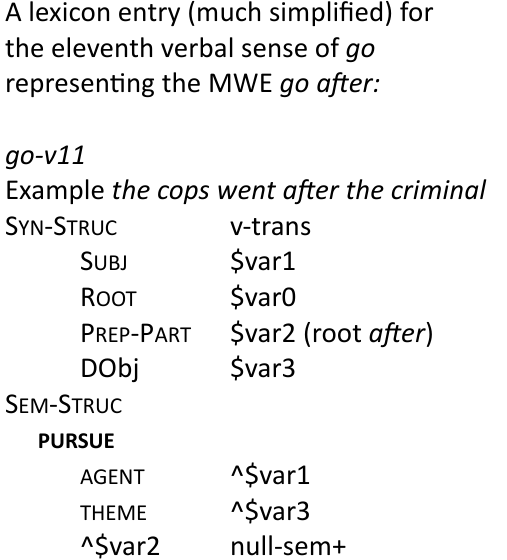}
    \end{framed}
    \caption{An MWE (phrasal verb) entry.}
    \label{fig:phrasal_verb_entry}
    \vspace{-5pt}
\end{figure}
The goal of our experimental set-up is to expand the system’s static knowledge resources by automatically learning other ways of expressing in English the meanings of verbal senses, such as \textit{employ-v3}. This task cannot be carried out with confidence by using thesauri or WordNets because their offerings are too imprecise and do not conform to the "seed" word either semantically or (especially for verbs) syntactically. \footnote{For a detailed discussion of why existing dictionaries, thesauri and wordnets cannot be used as resources for automating knowledge acquisition, \iffalse\fi\iftrue see Chapter 3 of McShane et al. forthcoming. \fi}
The general LLM-supported learning procedure we used in this experiment consists of five steps described below. 

\begin{enumerate}[label=\textbf{Step \arabic*:},leftmargin=15pt,listparindent=-\leftmargin, align=left, itemsep=-3pt]
\item Select a seed verb sense from the lexicon. Using the \textsc{sem-struc} zone of this sense, construct a semantically correct statement in the ontological metalanguage using ontological constraints on the verb sense’s case roles listed in the entry or in the ontology. For example, the \textsc{agent} of \textsc{hire} (which is the ontological concept expressing the meaning of \textsc{employ-v3}) is constrained to \textsc{managerial-role}, while its \textsc{theme} is constrained to \textsc{human}\footnote{This is a simplification, as the lexicon and ontology feature multiple layers of constraints. We use the grain size of description sufficient to explain the experimental procedure.}.  So, the resulting general template will be \textsc{hire} (\textsc{managerial-role}, \textsc{human}). 
\item Use the above template to automatically construct a set of sample generation-oriented meaning representations (GMRs) that substitute lexicon entries whose meanings directly reference the constraints in the template. An example of such a GMR would be \textit{employ-v3} (\textit{manager-1}, \textit{actor-1}). It conforms to the semantic constraints on \textsc{hire} because the meaning of \textit{manager-1} is \textsc{managerial-role}, and that of \textit{actor-1} is \textsc{human}. In our experiment, these GMR sets were constrained in size to reduce the compute requirements.
\item Use the text generator module of our agent system to generate English sentences from the above GMRs. The resulting set of sentences is our best bet to convey the meaning of the seed verb to an analogical reasoning engine such as an LLM (without much additional training LLMs will not be able to operate over formal meaning representations such as the GMRs). This method of explaining lexical meaning is what people usually use, though people, of course, can both use analogical reasoning and invoke world knowledge in interpreting novel lexical material.
\item Use an LLM to generate a list of multiword expressions (MWEs – typically, verbs with postpositions, e.g., \textit{bring in}) semantically equivalent to the seed verb sense whose contextual meaning is illustrated by the set of sentences generated in step 3. Details of this process form the core of this paper’s content and are detailed in Section \ref{Sec:2} below.
\item Construct a new lexical sense for each of the MWEs output from Step 4.  This is done by "cloning" the sense of the seed verb (see Figure \ref{fig:phrasal-verb-entry} for an example), adding the requirement for the presence of the lexical material other than the main verb to the \textsc{syn-struc} zone of the newly learned entry (the postposition \textit{after} in the example) and modifying the \textsc{sem-struc} zone of the entry by including the instruction that this "extra" lexical material does not carry any semantic weight when appearing in this MWE (marked as null-sem+). A sampling of sentences generated for validation purposes in Step 4 is listed in the EXAMPLE zone of the new entry. (As a side effect, a sample of the sentences generated in Step 3 is added to the EXAMPLE zone of the entry of the seed verb sense.) The new lexical senses are marked as "learned". The agent is still free to use them but may consider them less reliable when scoring TMR candidates during future agent processing. The intention is for a knowledge engineer at some point to inspect – and possibly edit – these learned senses before removing the "learned". marking. The level of human involvement in this kind of validation incurs significantly lower cognitive loads than manual lexicon acquisition, thus making the entire process less expensive and much more efficient.
\end{enumerate}

\section*{Adapting LLMs for Conceptual Learning}
\label{Sec:2}
The first issue in adapting LLMs for use in conceptual learning for cognitive agents is devising the most appropriate prompting architecture. Prompts serve as adjustable parameters that fine-tune the language models to yield desired outputs for specific tasks. In plain words, it is not enough to supply the LLMs with the output of the previous stage of the learning process, even if measures are taken to represent that output in natural language rather than a formal metalanguage that LLMs will not be able to interpret. We need to tell the LLMs what we want them to do. Many strategies have been proposed for designing LLM prompts and prompting architectures \cite{zhou2022large}. It has been shown that composing a single optimal prompt can be both challenging and ineffective, especially considering that LLM behavior is unpredictable. One proposed improvement is the “chain of thought” prompting approach \cite{wei2022chain}, which involves simplifying complex prompts by breaking them down into a series of manageable intermediate steps. This technique also allows for dynamic adaptation of process flows based on the outputs at intermediate stages \cite{yao2022react} and risk mitigation through validation. 

\begin{figure}[h]
    \centering
    \includegraphics[width=1\linewidth]{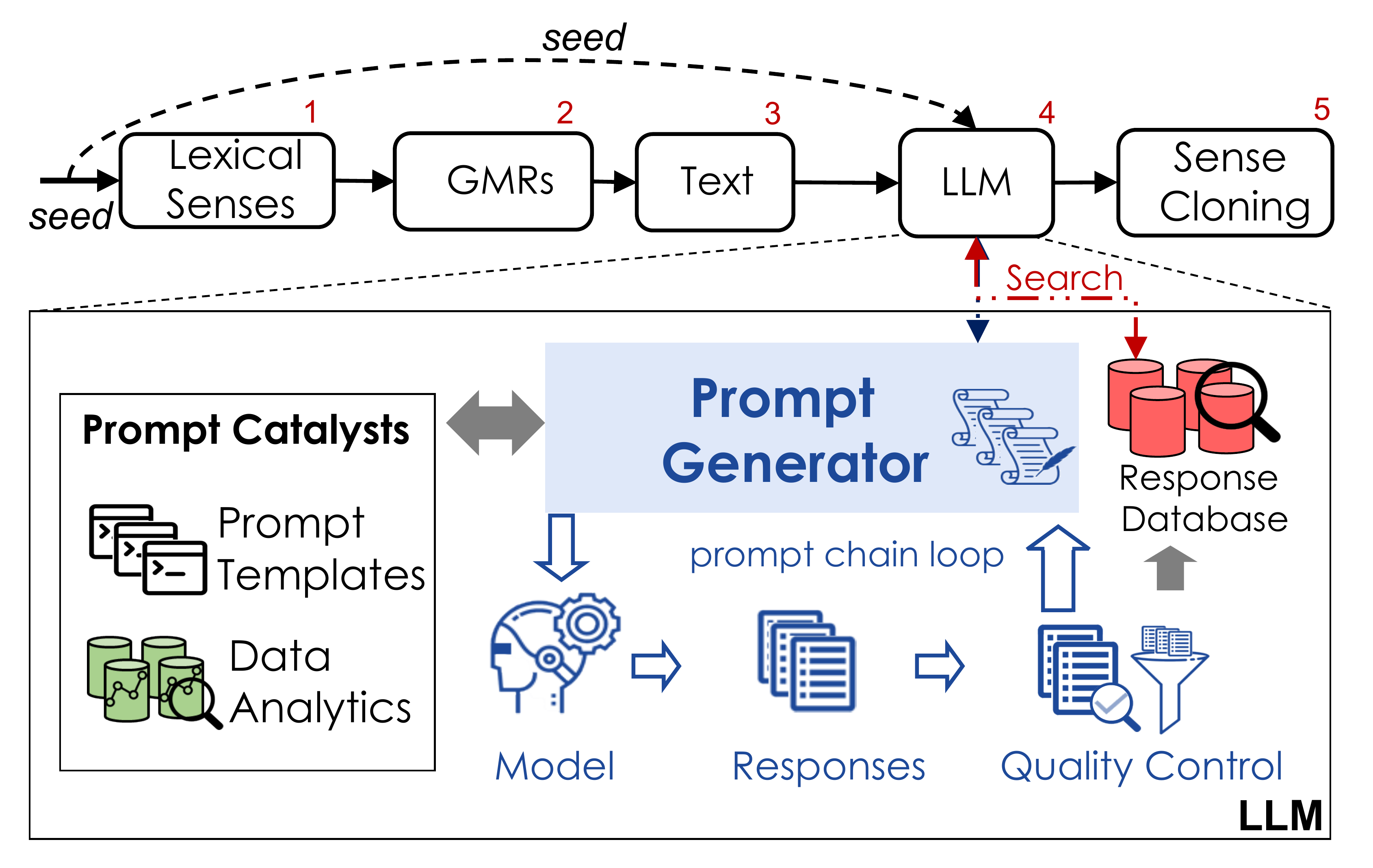}
    \caption{A schematic view of the LLM-Supported knowledge acquisition framework for language-endowed cognitive agents. The process is “seeded” with a word sense as input. Next, the knowledge-based language processor generates a set of semantically correct seed sentences containing that word sense. The seeds are incorporated into a chain of prompts to an LLM whose responses (in this experiment, MWEs and sample sentences containing them) are validated by asking the LLM itself to assess whether the semantics of the results it produces correspond to that of the seed word in the sense illustrated by the seed sentences. An alternative validation method substitutes sentences attested in the COCA corpus for sample sentences generated by LLMs.}
    \label{fig:overview}
\end{figure}

Constructing single monolithic prompts with placeholders for the seed verb sense and sentences generated in step 3 might seem to offer a simpler approach to prompting the LLM. For example, a prompt could be as simple as “In the context of the {text}, generate phrasal verbs that could replace the {seed}. Also, provide a contextually accurate example for the substituted phrasal verb.” However, when tested using both GPT-3.5 and GPT-4 \cite{radford2019language}, this approach proved ineffective: LLMs returned inadequate results. This prompt template proved too complex, as it contained multiple instructions and potential ambiguity due to insufficient detail. To improve the quality of LLM output, we enhanced the prompting architecture with “prompt catalysts” (see Figure \ref{fig:overview}): a) using prompt template sequences and b) embedding analytics obtained from reliable data sources \cite{santu2023teler}.

Our approach to prompt engineering facilitates conveying the semantics of the seed verb sense to an analogical reasoning engine, such as an LLM, as precisely as possible without the use of formal metalanguage. For the experiment we describe, we designed a prompt template chain (Figure \ref{fig:overview}) consisting of a base generic prompt template followed by three templates with placeholders for specific input data to support to support three substeps of Step 4 in the overall learning process: 1) a template for generating synonymous MWEs, 2) a template for generating sentences with these MWEs needed for validation and 3) a template for validation proper. The template chain we constructed effectively follows the chain of thought prompting approach (Wei 2022). Each prompt in the chain is sent as input to the LLM (we used the GPT-3.5 APIs from Open AI for this experiment). The LLM response to each prompt is then embedded in the subsequent prompts in the chain, which is necessary because LLMs lack memory of their prior processing.

\begin{figure*}[!t]
    \centering
    \includegraphics[width=\textwidth]{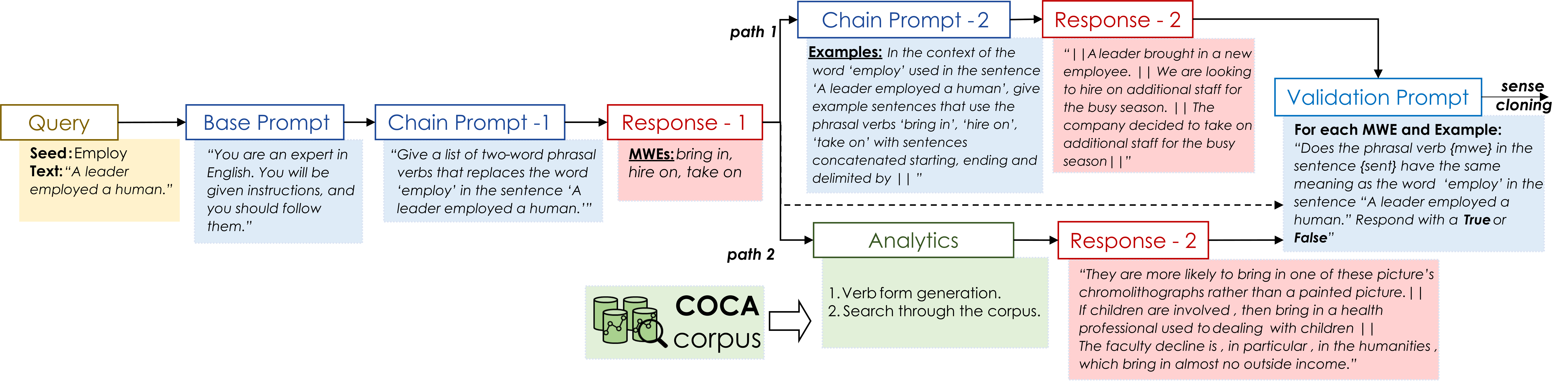}
    \caption{An example of the learner operation illustrating the generation of MWEs synonymous with the seed verb and two methods of generating sample sentences for use in the validation step: Path 1 illustrates the use of LLMs while Path 2 illustrates the use of the COCA corpus for this purpose. In every iteration of the experiment, only one of the two paths is selected for sentence generation, with its results then forwarded to the validation step. A different LLM  may be used for each of the three applications of LLMs in this process, should this option prove beneficial.}
    \label{fig:prompt}
\end{figure*}

The primary function of the base prompt is to present a particular task to the LLM and explain its role in completing that task. The base prompt does not include any placeholders for data. For instance, in the current experiment, the base prompt involves a simple role assignment, as illustrated in Figure \ref{fig:prompt}. 

The MWE generation substep of Step 4 of the learning process requires the seed verb and the set of sentences  (the output of Stage 3 of the learning process) as placeholders for the prompt template. At the validation substep, the LLM is asked to assess the quality of its own results from the generation substep. LLMs do this essentially by comparing the “gold standard” sentences generated in Step 3 of the learning process to illustrate the meaning of the seed verb sense against sentences containing the MWEs it generated at the previous substep. So, following the chain of thought method, the prompt for the validation substep incorporates a) the content of all the upstream prompts, b) the content of the response from the MWE generation substep, and c) a set of sentences containing each of the MWEs in this response. So, before validation can be triggered, an intermediate substep devoted to deriving the set of sentences in c) above is required.

How should we go about generating this set of sentences? We implemented two approaches (labeled as “paths” in Figure \ref{fig:prompt}): a) using the LLM itself to generate the set and or b) using data analytics to conduct a search in a text corpus for sentences containing the MWEs suggested by the LLM at the MWE generation substep. The LLM approach, as always, involved prompting the model with the cumulative prompt incorporating the latest response and the upstream prompts. The analytics approach used the COCA corpus (Davies 2008) and seeded the search with the set of all the morphological forms of the main verb in the MWE. In our experimentation, results from one or the other of the methods were used intermittently as inputs to the downstream validation step. 

In the LLM path (Path 1), additional filtering proved to be necessary to filter out irrelevant sentences from LLM-generated responses, specifically, sentences the LLM generated to shape its response as a dialog turn, as it was trained to simulate a behavior of a conversational companion. Figure \ref{fig:prompt} provides an illustration. For our purposes, the LLM response should ideally consist exclusively of sample sentence candidates separated by the '$||$' tag. However, LLMs (GPTs 3.5 in this case) generates extraneous material. To filter out such material, our system looks for the content separated by a specific tag (‘$||$’). In most cases, the elements of the response that do not begin with a tag carry extraneous material.

\begin{figure}[h]
    \centering
    \begin{framed}
    \raggedright
       \textbf{WORD:} measure\\
       \textbf{TEXT:} \textit{An actor measured a matter}.\\
       \textbf{LLM RESPONSE:}\\
       \textit{"\textcolor{red}{\sout{I apologize for the confusion. Here are several example sentences illustrating the use of the phrasal verb ‘take stock’: }}$||$ Let’s take stock of our inventory before placing the order. $||$ After a long day at work, I like to take stock of my accomplishments$||$”}\\
       \textbf{FILTERED RESPONSE (candidates):}\\
       \begin{itemize}
           \item Let’s take stock of our inventory before placing the order.
           \item After a long day at work, I like to take stock of my accomplishments.
       \end{itemize}
          \end{framed}
    \caption{An example showing filtering applied to LLM response in path 1. Here the excessive text that is in red was removed and the sentences delimited by ‘$||$’ }
    \label{fig:example1}
    \vspace{-10pt}
\end{figure}

In the analytics path (Path 2), the COCA search is preceded by using the NLTK libraries to generate all morphological forms of the main verb in the MWEs – e.g., {\textit{look up, looks up, looking up, looked up}}. Note that for both of the above methods, we have the ability to generate or select sentences where the MWE components are discontinuous. So, for example, the sentence "The puppy managed to run, suddenly, off towards the park could" be found for the MWE "run off."

\begin{figure}[h]
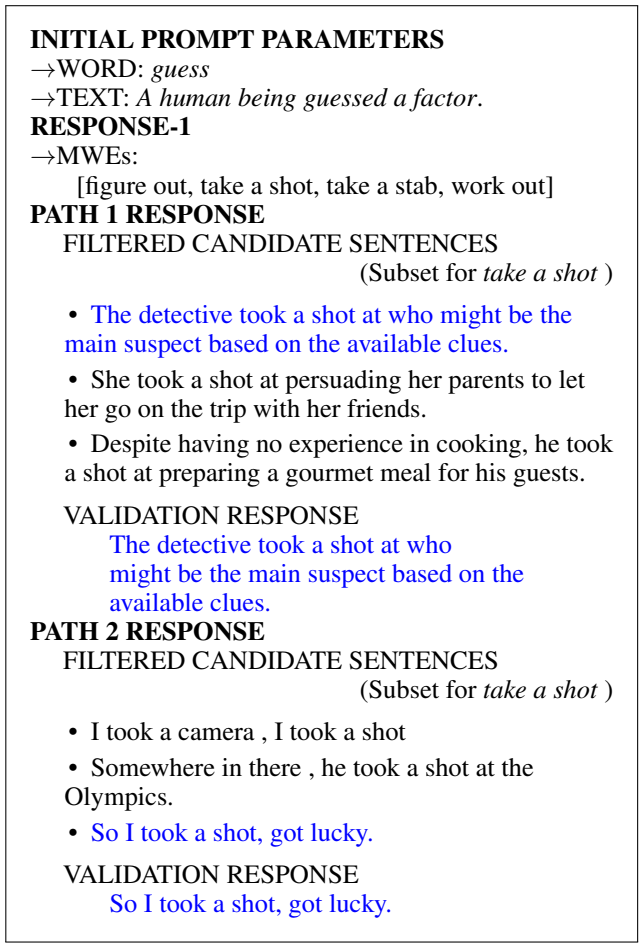

    \centering
    \begin{framed}
    \raggedright
    \textbf{INITIAL PROMPT PARAMETERS}\\
       $\rightarrow$WORD: \textit{guess}\\
       $\rightarrow$TEXT: \textit{A human being guessed a factor}.\\
       \textbf{RESPONSE-1}\\
       $\rightarrow$MWEs:\\\phantom{.}\phantom{.}\phantom{.}\phantom{.}\phantom{.}\phantom{.}\phantom{.}[figure out, take a shot, take a stab, work out]\\
       \textbf{PATH 1 RESPONSE}\\
       \phantom{.}\phantom{.}\phantom{.}\phantom{.} FILTERED CANDIDATE SENTENCES \\
       \hfill(Subset for \textit{take a shot} )\\
       \begin{itemize}
  
  \addtolength{\itemindent}{1em}  
           \item \textcolor{blue}{The detective took a shot at who might be the main suspect based on the available clues.}
           \item She took a shot at persuading her parents to let her go on the trip with her friends. 
           \item Despite having no experience in cooking, he took a shot at preparing a gourmet meal for his guests.
       \end{itemize}
       \phantom{.}\phantom{.}\phantom{.}\phantom{.} VALIDATION RESPONSE \\
       \setlength{\parindent}{30pt}\textcolor{blue}{The detective took a shot at who\\
       \setlength{\parindent}{30pt}might be the main suspect based on the\\
       \setlength{\parindent}{30pt}available clues. }\\

        \setlength{\parindent}{0pt}\textbf{PATH 2 RESPONSE}\\
       \phantom{.}\phantom{.}\phantom{.}\phantom{.} FILTERED CANDIDATE SENTENCES \\
       \hfill(Subset for \textit{take a shot} )\\
       \begin{itemize}
  
  \addtolength{\itemindent}{1em}  
        \item I took a camera , I took a shot
        \item Somewhere in there , he took a shot at the Olympics.
           \item \textcolor{blue}{So I took a shot, got lucky.}
           
       \end{itemize}
       \phantom{.}\phantom{.}\phantom{.}\phantom{.} VALIDATION RESPONSE \\
       \setlength{\parindent}{30pt}\textcolor{blue}{So I took a shot, got lucky.}

          \end{framed}
    \caption{Example 1 showing MWEs generated for the seed word \textit{guess} and the context text input from step 3. The example shows candidate sentences for the \textit{take a sho}t MWE obtained from paths 1 and 2 and their corresponding validation responses. }
    \label{fig:eg1}
    \vspace{-5pt}
\end{figure}

\begin{figure}[!t]
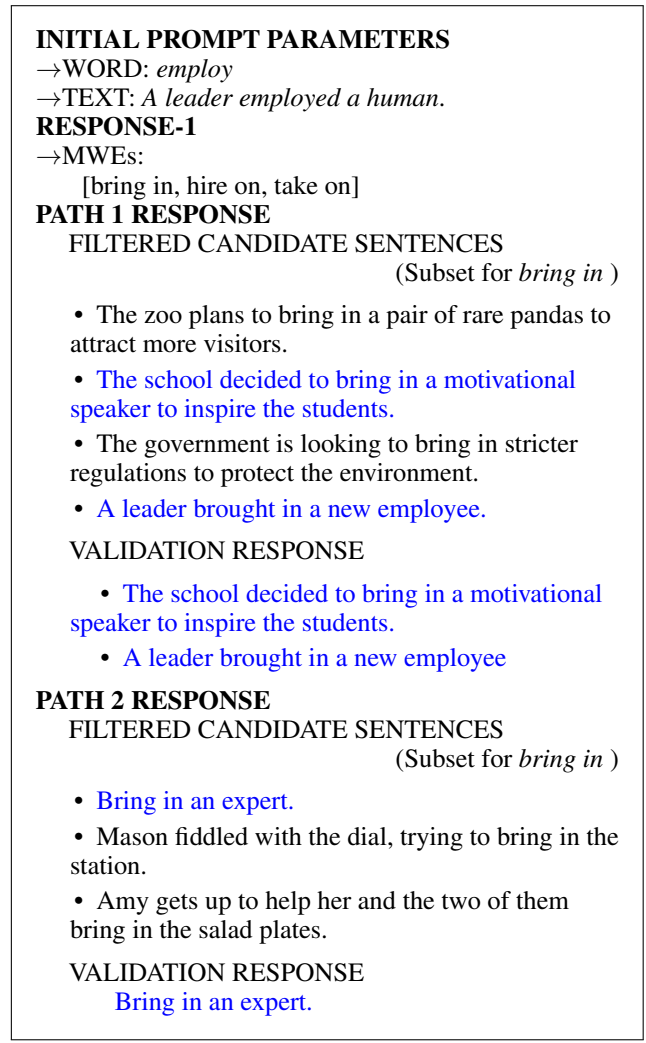

    \centering
    \begin{framed}
    \raggedright
    \textbf{INITIAL PROMPT PARAMETERS}\\
       $\rightarrow$WORD: \textit{employ}\\
       $\rightarrow$TEXT: \textit{A leader employed a human}.\\
       \textbf{RESPONSE-1}\\
       $\rightarrow$MWEs:\\\phantom{.}\phantom{.}\phantom{.}\phantom{.}\phantom{.}\phantom{.}\phantom{.}[bring in, hire on, take on]\\
       \textbf{PATH 1 RESPONSE}\\
       \phantom{.}\phantom{.}\phantom{.}\phantom{.} FILTERED CANDIDATE SENTENCES \\
       \hfill(Subset for \textit{bring in} )\\
       \begin{itemize}
  
  \addtolength{\itemindent}{1em}  
            \item The zoo plans to bring in a pair of rare pandas to attract more visitors. 
            
           \item \textcolor{blue}{The school decided to bring in a motivational speaker to inspire the students.}
           \item The government is looking to bring in stricter regulations to protect the environment.
           \item \textcolor{blue}{A leader brought in a new employee.}
       \end{itemize}
       \phantom{.}\phantom{.}\phantom{.}\phantom{.} VALIDATION RESPONSE \\
        \begin{itemize}
          \addtolength{\itemindent}{2em} 
            \item \textcolor{blue}{The school decided to bring in a motivational speaker to inspire the students.}
             \item \textcolor{blue}{A leader brought in a new employee}
            
        \end{itemize}
        \setlength{\parindent}{0pt}\textbf{PATH 2 RESPONSE}\\
       \phantom{.}\phantom{.}\phantom{.}\phantom{.} FILTERED CANDIDATE SENTENCES \\
       \hfill(Subset for \textit{bring in} )\\
       \begin{itemize}
  
  \addtolength{\itemindent}{1em} 
  \item \textcolor{blue}{Bring in an expert.}
        \item Mason fiddled with the dial, trying to bring in the station.
        \item Amy gets up to help her and the two of them bring in the salad plates.
       \end{itemize}
       \phantom{.}\phantom{.}\phantom{.}\phantom{.} VALIDATION RESPONSE \\
       \setlength{\parindent}{30pt}\textcolor{blue}{Bring in an expert.}
          \end{framed}
    \caption{Example 2 showing MWEs generated for the seed word \textit{employ} and the context text input from step 3. The example shows candidate sentences for the \textit{bring in} MWE obtained from paths 1 and 2 and their corresponding validation responses. }
    \label{fig:eg2}
    \vspace{-5pt}
\end{figure}

 The filtering process described above is part of the sentence generation substep of Step 4 of the overall process and differs from the validation substep. The former targets irrelevant sentences while the latter carries out a more semantically meaningful task: it filters out sentences that contain the MWEs used in a sense different from that of the seed verb sense. Figures \ref{fig:eg1} and \ref{fig:eg2} illustrate some results of validating the results of the operation of each of the two sentence generation methods (Path 1 and Path 2). Thus, in Figure \ref{fig:eg2} the MWE \textit{bring in} in the sense illustrated by the sentence "The government is looking to bring in stricter regulations to protect the environment" was not validated as carrying the same meaning as the seed verb sense (\textit{employ-v3}) whose meaning is illustrated by the seed text \textit{A leader employed a human} generated at Step 3 of the overall learning process. As a result, the MWE \textit{bring in} was eliminated from the list of candidates.

Once the automatic validation of the newly learned set of MWEs is completed, the learning process advances to Step 5 of the overall learning algorithm, as depicted in Figure \ref{fig:overview}.
\section*{Discussion}

We view the experiment described in this paper as an early step in a multifaceted R\&D effort on using LLMs to support automatic learning by language-endowed AI agents. We intend this learning environment to integrate all and all and any methods and resources that can be shown to contribute to the task. The experiment reported here demonstrates a method that integrates LLMs and data analytics with knowledge-oriented methods and resources in a truly hybrid architecture. 

The system developed for the experiment we describe does not yet make use of an important capability our agent systems possess – the ability to extract and represent in an ontologically-motivated metalanguage the set of semantic and discourse/pragmatic meanings of a text. We have experimented with using this capability for learning in the past \iftrue(e.g., English and Nirenburg 2010)\fi, and we intend to incorporate it in the learning environment described here. This program of work may seem to aim at autonomous learning. But in fact we view our learning environment as an orthotic system \iftrue(Nirenburg 2017) \fi that expects human participation in several roles, notably as an instructor in a dialog set-up and as a knowledge engineer responsible for tuning up and maintaining the system. Indeed, when our system is deployed and starts regular operation, the MWE (and, later, other) lexical senses added to the lexicon as a result of the learning process we describe here can – and will – be vetted by knowledge engineers.

Our team's experimentation on the integration of knowledge-based methods with LLMs is not restricted to learning applications. Our agents use LLMs, for example, at a final step of text generation to select the most contextually appropriate of the candidate English sentences generated by the agent’s text generator from formal representations of the meaning of the message that must be conveyed. While all of the options from which the LLM is asked to choose are semantically and syntactically correct, the LLM has been shown to select the most appropriate one stylistically and contextually, thus obviating the need to develop a conceptual system module for this purpose.

\section*{Future Work}

At the time of writing, we are testing the learning process we described on the content of our agent’s lexicon. At the symposium, we will present the results of an evaluation of the utility of this method when applied to all 1,153 senses of transitive verbs in the our system's current lexicon. We are also planning to extend the learning environment to address other types of lexical material, such as learning single-word (not MWE) true synonyms for transitive verbs, intransitive verbs and other parts of speech. Learning new and improving existing ontological concepts will be tackled next.

The overall learning process will itself be enhanced and improved. For example, in creating GMRs in Step 2 of the process we have not yet used ontological descendants of the concepts constraining the meanings of the seed verb sense’s subject and direct object. We intend to experiment with this option in the immediate future. 

Other planned extensions include using the learning environment to carry out “inverted” on-the-fly learning as an approach to treating unexpected input during the agent’s regular operation: semantically true paraphrases for an unknown verb in a textual input can be generated using the method presented in this paper, in hopes that at least some of the paraphrases the system will find are already attested in the lexicon. For example, if the word \textit{buff} is not in the system's lexicon, but our LLM-supported learning process suggests an existing sense of \textit{polish} as having the same meaning, this will yield a double bonus of a) the system succeeding in generating a meaning representation for the sentence containing \textit{buff} using the content of the lexicon entry for \textit{polish} and b) the side effect of creating a lexicon entry for \textit{buff} on the fly. This kind of learning is known as opportunistic \iftrue(e.g., McShane et al., forthcoming, Chapter 6) \fi. 

The same operation can also be implemented as part of deliberate learning, a mode in which the agent does not carry out any other tasks.  This requires a preliminary step of detecting lexicon units not present in the lexicon that are candidates for learning. Finally, we intend to investigate how to go beyond synonymy in this approach to learning and learn near-synonyms (plesionyms) as well as hypo- and hypernyms and other lexical units related to a seed lexicon sense. 

In our current implementation, we used pre-trained LLMs. Training LLMs for specific tasks related to learning by cognitive agents is an additional avenue for future development. The LLMs we use are trained on text. We intend to experiment with training them on a mixture of text and text meaning representations (TMRs) generated and used by our agent systems. The prerequisite for this is generating large stores of TMRs. This effort is at already underway in our research team.

\section*{Acknowledgments}
This research is supported in part by the Office of the Director of National Intelligence (ODNI), Intelligence Advanced Research Projects Activity (IARPA), via the HIATUS Program contract \# 2022-22072200001. 

The views and conclusions contained herein are those of the authors and should not be interpreted as necessarily representing the official policies, either expressed or implied, of ODNI, IARPA, or the U.S. Government. The U.S. Government is authorized to reproduce and distribute reprints for governmental purposes, notwithstanding any copyright annotation therein.

\bibliography{aaai23}

\end{document}